\documentclass[letterpaper]{article} 
\usepackage{aaai23}  
\usepackage{times}  
\usepackage{helvet}  
\usepackage{courier}  
\usepackage[hyphens]{url}  
\usepackage{graphicx} 
\usepackage{natbib}  
\usepackage{caption} 
\usepackage{algorithm}
\usepackage{algorithmic}

\usepackage{newfloat}
\usepackage{listings}
\DeclareCaptionStyle{ruled}{labelfont=normalfont,labelsep=colon,strut=off} 
\floatstyle{ruled}
\newfloat{listing}{tb}{lst}{}
\floatname{listing}{Listing}
\usepackage{subcaption}
\usepackage{amsmath}
\usepackage{amssymb}
\newtheorem{definition}{Definition}

\title{Improving Adversarial Robustness to Sensitivity and Invariance Attacks with Deep Metric Learning}
\author{
    Anaelia Ovalle\equalcontrib,
    Evan Czyzycki \equalcontrib, 
    Cho-Jui Hsieh
}
\affiliations{
    \textsuperscript{\rm 1} University of California, Los Angeles \\
    \{anaelia, eczy, chohsieh\}@cs.ucla.edu
}

\begin{document}

\maketitle

\begin{abstract}
Intentionally crafted adversarial samples have effectively exploited weaknesses in deep neural networks. A standard method in adversarial robustness assumes a framework to defend against samples crafted by minimally perturbing a sample such that its corresponding model output changes. These \textit{sensitivity} attacks exploit the model's sensitivity toward task-irrelevant features. Another form of adversarial sample can be crafted via \textit{invariance} attacks, which exploit the model underestimating the importance of relevant features. Previous literature has indicated a tradeoff in defending against both attack types within a strictly $\ell_p$ bounded defense. To promote robustness toward both types of attacks beyond Euclidean distance metrics, we use metric learning to frame adversarial regularization as an optimal transport problem. Our preliminary results indicate that regularizing over invariant perturbations in our framework improves both invariant and sensitivity defense.
\end{abstract}

\section{Introduction}
Adversarial robustness can be motivated by the canonical comparison between 2 images: a panda and a panda with an imperceptibly small perturbation that results in a completely different image classification \cite{Goodfellow2015ExplainingAH}. Such adversarial attacks exploit a model's sensitivity to features that it considers highly important to the learning task but are actually of little significance \cite{tramer2020fundamental}. However, a less studied class of adversarial samples exploits a model's invariance to relevant features. 

Ensuring safety in machine learning algorithms requires the field of adversarial robustness to be keen on examining new forms of attack and subsequent mitigation strategies. \citet{tramer2020fundamental} explore invariance attacks and observe a fundamental tradeoff in defending sensitivity and invariance attacks when considering the neural networks as presented in \citet{jacobsen2020excessive}. Furthermore, \citet{tramer2020fundamental} find that using common adversarial training frameworks that rely on $\ell_p$ perturbations to improve robustness toward sensitivity attacks necessarily worsens robustness toward invariance attacks. Augmenting training solely with these constraints in Euclidean space causes the model to become increasingly invariant towards task-relevant features.

Navigating this tradeoff has not been explored in non-Euclidean spaces. In this study, we propose an adversarial framework that deviates from creating perturbations under Euclidean norms. Instead, we frame robustness measures as an optimal transport problem to be solved via metric learning \cite{10.5555/3491440.3491739}. This approach allows us to produce adversarial samples anchored outside a Euclidean $\ell_p$ bounded ball, which allows us to regularize over both invariance and sensitivity attacks simultaneously.

\section{Sensitivity and Invariance Adversarial Attacks}
Let us consider a classification task with samples $(x, y) \in \mathbb{R}^d x \{1, ..., C\} \sim D$. Let us also consider a ground truth labeling oracle $\mathcal{O}: \mathbb{R}^d \rightarrow \{1, ..., C\}$.

\begin{definition}[Sensitivity Adversarial Example]
Given some classifier $f$,  and a correctly classified input $(s, y) \sim D$, an $\epsilon$-bound sensitivity adversarial example is an input $x^* \in \mathbb{R}^d$ such that:
\begin{enumerate}
    \item $f(x^*) \neq f(x)$.
    \item $||x^* - x|| \leq \epsilon$.
\end{enumerate}
\end{definition}

\begin{definition}[Invariance Adversarial Example]

Given some classifier $f$,  and a correctly classified input $(s, y) \sim D$, an $\epsilon$-bound invariance adversarial example is an input $x^* \in \mathbb{R}^d$ such that:
\begin{enumerate}
    \item $f(x^*) = f(x)$.
    \item $\mathcal{O}(x^*) \neq \mathcal{O}(x)$ and $\mathcal{O}(x^*) \neq \bot$.
    \item $||x^* - x|| \leq \epsilon$.
\end{enumerate}
\end{definition}

Note that the above formulation and definitions mirror those found in \citet{tramer2020fundamental}. A significant assumption required for Definition 1 is that for all $x$ and associated perturbations $x^*$, if $||x^* - x|| \leq \epsilon$, then $\mathcal{O}(x^*) = \mathcal{O}(x)$. Informally, perturbations of magnitude less than $\epsilon$ preserve the oracle's labelling. As shown in \citet{tramer2020fundamental}, it is precisely the violation of this assumption that results in a fundamental tradeoff between robustness toward these two types of adversarial attacks.

\begin{table*}[ht]
\centering
\small
\begin{tabular}{l|l|l|l} 
\hline
\textbf{Model}                                          & \textbf{Original Image} & \textbf{Sensitivity Attack (FGSM)}                                  & \textbf{Invariance Attack}                           \\ 
\hline
Adversarial Training with FGSM (Baseline)                             & 99.02          &  98.95 \           & 85.67        \\ 
\hline
Baseline + MLS & 99.38          & 99.17                                          & 82.49                                         \\ 
\hline
Baseline + MLS + MLI   & 99.09          & \textbf{98.98} & \textbf{87.80}   \\
\hline
\end{tabular}
\caption{Accuracy over MNIST data. Values are averaged over three randomly seeded runs. Regularizing with angular triplet loss for both sensitivity and invariance attacks improves accuracy over invariance samples with minimal impact on sensitivity accuracy. MLS and MLI indicate the usage of a metric learning norm with sensitivity or invariance samples, respectively.}
\label{tab:results}
\vspace{-5mm}
\end{table*}

\section{Adversarial Metric Learning Framework}
In our training we transform embeddings to angular space and use metric learning to implicitly learn a distance measure. This results in a more flexible adversarial optimization framework as shown in \citet{duan2018deep}. Whereas this study investigates the use of metric learning in traditional adversarial defense against sensitivity attacks, however, our study explores using this framework to defend against invariance attacks as well.

Our loss function, $L_{t}$, is defined as the classic triplet loss found in \cite{mao2019metric}, which creates a fixed margin between the differences in the anchor sample and positive and negative examples respectively. We define the distance, $D(\cdot)$, as the angular distance between two samples in order to encode the information in the angular metric space.

\vspace{-5mm}


\begin{equation}
D(h(\textbf{x}_{a}^{(i)}), h(\textbf{x}_{p, n}^{(i )})) = 1 - \frac{\vert{h(\textbf{x}_{a}^{(i)})}\cdot{h(\textbf{x}_{p,n}^{(j)}})\vert} {\Vert{h(\textbf{x}_{a}^{(i)})}\Vert_{2}\Vert{h(\textbf{x}_{p,n}^{(j)}})\Vert}
\end{equation}




We derive an adversarial training framework using the below loss term by considering the anchor sample to be a natural image $x_a$, a positive example to be a perturbed image $x_p$, and a negative sample to be an image $x_n$ from a different class. We construct our loss function by including a sensitivity triplet loss regularization term, an invariance triplet loss regularization term, and a feature norm. This forces adversarial and natural samples closer together in learned space.

\vspace{-5mm}
\begin{align}
\begin{split}
    L_{all} &= \sum_{i}^{N} L_{ce}(f(\textbf{x}_{a}^{(i)}), y^{(i)}) \\
    &+\lambda_{1}L_{t,sa}((h(\textbf{x}_{a}^{(i)}) ,h(\textbf{x}_p^{'(i)}),h(\textbf{x}_{n}^{(i)})) \\
    &+ \lambda_{2}L_{t,ia}((h(\textbf{x}_{a}^{(i)}) ,h(\textbf{x}_p^{"(i)}),h(\textbf{x}_{n}^{(i)})) \\
    &+ \lambda_{3}L_{norm}
\end{split}
\end{align}
where
\begin{align*}
    L_{norm} &= \Vert h(\textbf{x}_{a}^{(i)}) \Vert_{2} +\Vert h(\textbf{x}_{p}^{'(i)}) \Vert_{2} \\
    &+ \Vert h(\textbf{x}_{p}^{"(i)}) \Vert_{2} + \Vert h(\textbf{x}_{n}^{(i)}) \Vert_{2}
\end{align*}

$L_{ce}$ is cross entropy loss, $L_{t,sa}$ uses sensitivity attacks for positive class, $L_{t,ia}$ uses invariance attacks for positive class, $x_{p}^{'}$ and $x_{p}^{"}$ are sensitivity and invariance perturbed samples respectively, and $\lambda_1,\lambda_2,\lambda_3$ are coefficients $\in \mathbb{R}^{+}$.

\section{Experiments and Discussion}
In our experiments we generate sensitivity attacks using FGSM \cite{Goodfellow2015ExplainingAH} and invariance attacks using the method described in \citet{tramer2020fundamental}. We generate a single sensitivity attack and a single invariance attack for each sample in the MNIST dataset.

Our results in Table \ref{tab:results} indicate that regularizing with sensitivity and invariance attacks using an angular triplet loss can improve performance against invariance attacks with minimal loss in accuracy over sensitivity attacks. Our model trained with sensitivity and invariance regularization outperforms the adversarial baseline which uses $\ell_p$-bound norms. 

\begin{figure}[htbp]
  \begin{subfigure}[b]{0.23\textwidth}
    \includegraphics[width=\columnwidth]{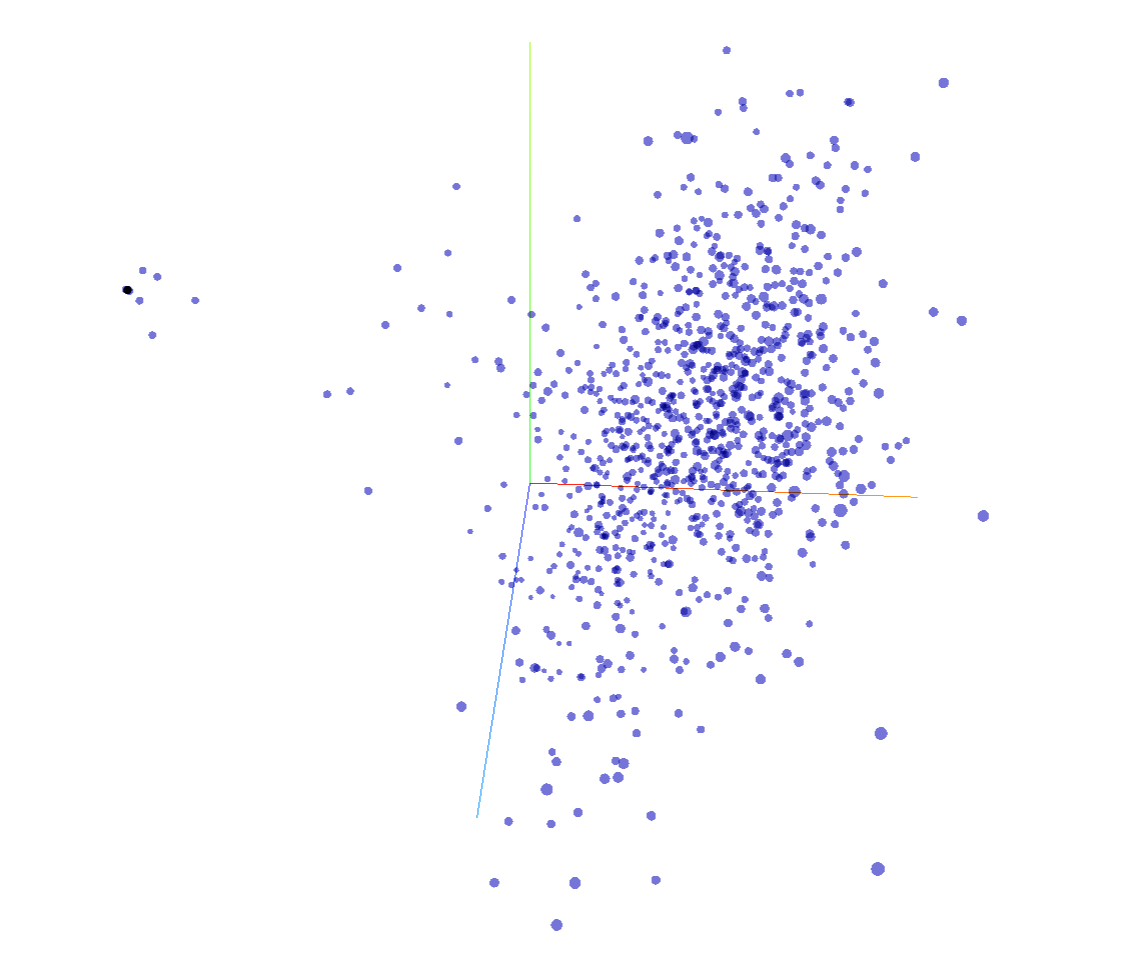}
    \caption{}
    \label{fig:pca-noreg}
  \end{subfigure}
  \hfill
  \begin{subfigure}[b]{0.23\textwidth}
    \includegraphics[width=\columnwidth]{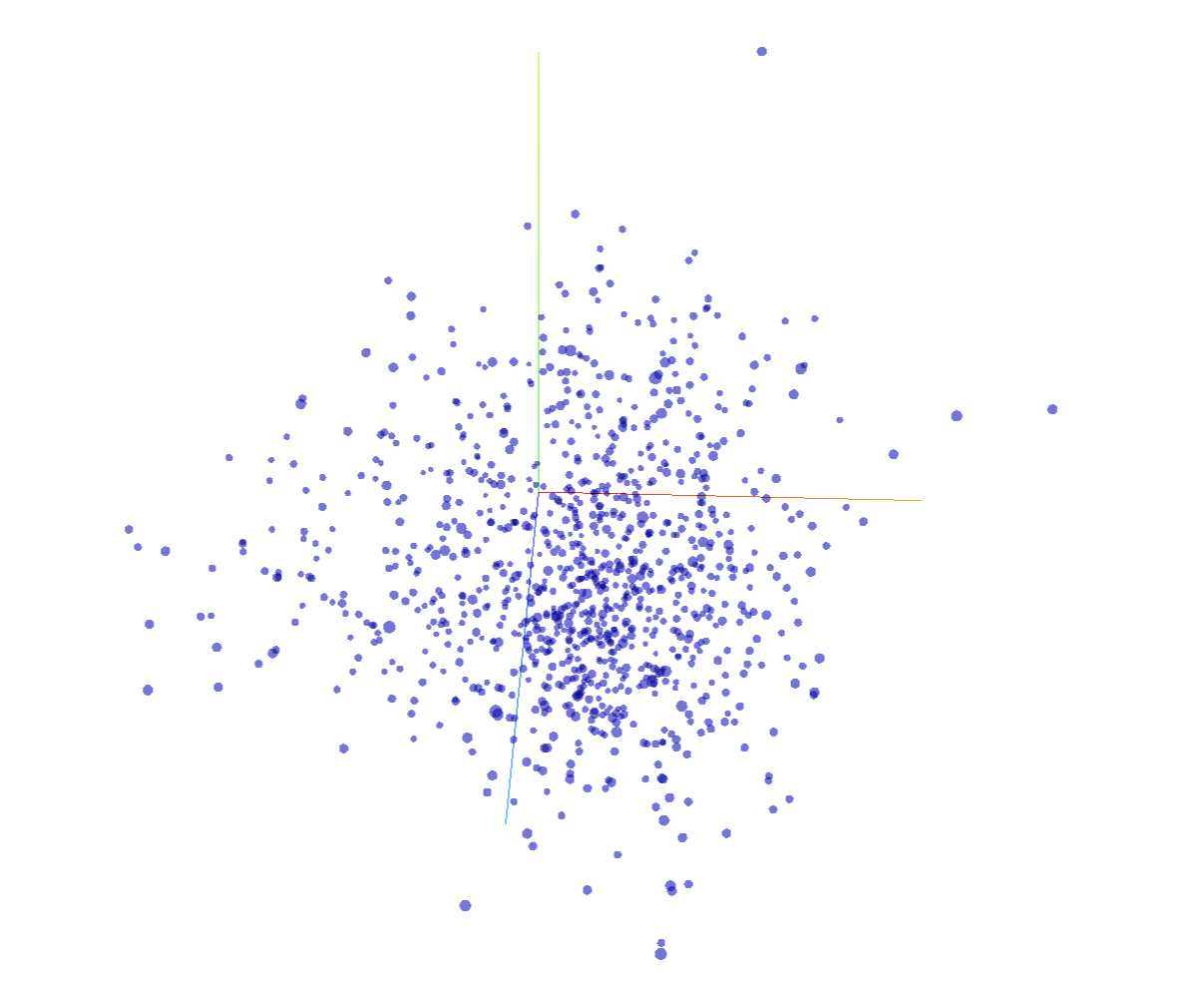}
    \caption{}
    \label{fig:pca-reg}
  \end{subfigure}
  \caption{(a) PCA projection of FGSM when the invariance regularizer not is added. (b) PCA projection of FGSM when the invariance regularizer is added}
\end{figure}

After each model is trained, we extract the penultimate layer and examine the learned embedding space with PCA. Figure \ref{fig:pca-noreg} shows the  distribution of adversarial images without invariance regularization. In comparison, when the invariance triplet regularizer is added, it is shown to be more tightly grouped and circular (\ref{fig:pca-reg}). This indicates our model's ability to better identify perturbed samples because they're grouped together. For future work, we plan to expand this analysis to more datasets and adversarial attacks.



\vspace{-5mm}
\bibliography{aaai23}

\begin{thebibliography}{6}
\providecommand{\natexlab}[1]{#1}

\bibitem[{Duan et~al.(2018)Duan, Zheng, Lin, Lu, and Zhou}]{duan2018deep}
Duan, Y.; Zheng, W.; Lin, X.; Lu, J.; and Zhou, J. 2018.
\newblock Deep adversarial metric learning.
\newblock In \emph{Proceedings of the IEEE Conference on Computer Vision and
  Pattern Recognition}, 2780--2789.

\bibitem[{Goodfellow, Shlens, and Szegedy(2015)}]{Goodfellow2015ExplainingAH}
Goodfellow, I.~J.; Shlens, J.; and Szegedy, C. 2015.
\newblock Explaining and Harnessing Adversarial Examples.
\newblock \emph{CoRR}, abs/1412.6572.

\bibitem[{Jacobsen et~al.(2020)Jacobsen, Behrmann, Zemel, and
  Bethge}]{jacobsen2020excessive}
Jacobsen, J.-H.; Behrmann, J.; Zemel, R.; and Bethge, M. 2020.
\newblock Excessive Invariance Causes Adversarial Vulnerability.
\newblock arXiv:1811.00401.

\bibitem[{Kerdoncuff, Emonet, and Sebban(2021)}]{10.5555/3491440.3491739}
Kerdoncuff, T.; Emonet, R.; and Sebban, M. 2021.
\newblock Metric Learning in Optimal Transport for Domain Adaptation.
\newblock In \emph{Proceedings of the Twenty-Ninth International Joint
  Conference on Artificial Intelligence}, IJCAI'20.
\newblock ISBN 9780999241165.

\bibitem[{Mao et~al.(2019)Mao, Zhong, Yang, Vondrick, and Ray}]{mao2019metric}
Mao, C.; Zhong, Z.; Yang, J.; Vondrick, C.; and Ray, B. 2019.
\newblock Metric Learning for Adversarial Robustness.
\newblock In Wallach, H.; Larochelle, H.; Beygelzimer, A.; d\textquotesingle
  Alch\'{e}-Buc, F.; Fox, E.; and Garnett, R., eds., \emph{Advances in Neural
  Information Processing Systems}, volume~32. Curran Associates, Inc.

\bibitem[{Tramèr et~al.(2020)Tramèr, Behrmann, Carlini, Papernot, and
  Jacobsen}]{tramer2020fundamental}
Tramèr, F.; Behrmann, J.; Carlini, N.; Papernot, N.; and Jacobsen, J.-H. 2020.
\newblock Fundamental Tradeoffs between Invariance and Sensitivity to
  Adversarial Perturbations.
\newblock arXiv:2002.04599.

\end{thebibliography}
\appendix

\end{document}